\pdfoutput=1

\documentclass[11pt]{article}

\usepackage{EMNLP2023}

\usepackage{times}
\usepackage{latexsym}

\usepackage[T1]{fontenc}

\usepackage[utf8]{inputenc}

\usepackage{microtype}

\usepackage{inconsolata}

\usepackage{graphicx}
\usepackage{float}
\usepackage{wrapfig}
\usepackage{caption}
\usepackage{amsmath}
\usepackage{booktabs}
\usepackage{amsfonts}  
\usepackage{bm}
\usepackage{makecell}

\title{Alif: Advancing Urdu Large Language Models via\\ Multilingual Synthetic Data Distillation}

\author{
  Muhammad Ali Shafique\textsuperscript{1},
  Kanwal Mehreen\textsuperscript{2},
  Muhammad Arham\textsuperscript{3} \\
  \textbf{Maaz Amjad}\textsuperscript{\textbf{4}},
  \textbf{Sabur Butt}\textsuperscript{\textbf{5}},
  \textbf{Hamza Farooq}\textsuperscript{\textbf{6}} \\
  \textsuperscript{1,3,6}Traversaal.ai\quad
  \textsuperscript{2}University of British Columbia\quad
  \textsuperscript{4}Texas Tech University \\
  \textsuperscript{5}Institute for the Future of Education, Tecnol\'ogico de Monterrey
}

\begin{document}
\maketitle
\maketitle
\begingroup
\renewcommand\thefootnote{}\footnote{Accepted to the EMNLP 2025 Workshop on Multilingual Representation Learning (MRL).}
\addtocounter{footnote}{-1}
\endgroup

\begin{abstract}
Developing a high-performing large language models (LLMs) for low-resource languages such as Urdu, present several challenges. These challenges include the scarcity of high-quality datasets, multilingual inconsistencies, and safety concerns. Existing multilingual LLMs often address these issues by translating large volumes of available data. However, such translations often lack quality and cultural nuance while also incurring significant costs for data curation and training. To address these issues, we propose Alif-1.0-8B-Instruct, a multilingual Urdu-English model, that tackles these challenges with a unique approach. We train the model on a high-quality, multilingual synthetic dataset (Urdu-Instruct), developed using a modified self-instruct technique. By using unique prompts and seed values for each task along with a global task pool, this dataset incorporates Urdu-native chain-of-thought based reasoning, bilingual translation, cultural relevance, and ethical safety alignments. This technique significantly enhances the comprehension of Alif-1.0-8B-Instruct model for Urdu-specific tasks. As a result, Alif-1.0-8B-Instruct, built upon the pretrained Llama-3.1-8B, demonstrates superior performance compared to Llama-3.1-8B-Instruct for Urdu specific-tasks. It also outperformed leading multilingual LLMs, including Mistral-7B-Instruct-v0.3, Qwen-2.5-7B-Instruct, and Cohere-Aya-Expanse-8B, all within a training budget of under \$100. Our results demonstrate that high-performance and low-resource language LLMs can be developed efficiently and culturally aligned using our modified self-instruct approach. All datasets, models, and code are publicly released\footnote{GitHub: \href{https://github.com/traversaal-ai/alif-urdu-llm}{github.com/traversaal-ai/alif-urdu-llm}}.
\end{abstract}

\section{Introduction}

The rapid advancement of LLMs~\citep{zhao2024surveylargelanguagemodels} has revolutionized natural language processing (NLP) across multiple languages and applications. However, a significant disparity persists between high-resource languages, such as English, and low-resource languages, such as Urdu. These disparities create technological barriers for billions of speakers of underrepresented languages, limiting their access to AI-driven tools and advancements. The inclusion of low-resource languages in LLM development is not merely a technical challenge but a crucial step toward fostering inclusive, globally accessible AI systems that cater to diverse linguistic communities.

Developing high-performing LLMs for low-resource languages presents several challenges, including the scarcity of high-quality datasets, multilingual inconsistencies, translation inaccuracies, reasoning limitations, and ethical concerns. A common approach to addressing these challenges relies on leveraging translated data from high-resource languages. However, translations often fail to capture regional knowledge and cultural nuances, leading to compromised language representation and ineffective communication in low-resource settings~\citep{Aharoni2019Massively, Conneau2020Unsupervised}. 

In the case of Urdu LLMs, additional factors contribute to their underperformance. Urdu’s linguistic complexity, including its unique alphabet, intricate grammar, syntax, and morphology, poses significant challenges in adapting NLP techniques developed for English. Furthermore, Urdu has borrowed extensively from regional languages such as Hindi, Punjabi, and Persian and is written in both the Perso-Arabic and Devanagari scripts, adding additional layers of complexity. While multilingual models exhibit some degree of understanding, their generation capabilities remain inadequate, particularly for languages with syntactic structures and writing systems distinct from English. Among these challenges, the lack of high-quality datasets stands out as a fundamental limitation. Current Urdu datasets are sparse, manually labeled, and contain only a few thousand instances—insufficient for training robust LLMs. This scarcity results from multiple factors, including limited digitization of Urdu literature, funding and infrastructure constraints, and the complexities of annotating Urdu text, which require linguistic expertise and standardized guidelines. Furthermore, translated data often fails to retain cultural nuances~\citep{alkhamissi2024investigating,ramaswamy2024geode}, such as idiomatic expressions and contextual meanings, thereby reducing a model’s ability to generate culturally relevant responses. Additionally, multilingual LLMs suffer from catastrophic forgetting, where training across multiple languages or modalities can degrade performance on certain language subsets unless carefully managed. The challenge of evaluation further complicates this issue~\citep{yu2022beyond}, as creating frameworks that fairly and accurately assess performance across diverse languages and cultures demands significant expertise and resources. These issues are particularly pronounced for South Asian low-resource languages like Urdu, which, despite its online presence, lacks the research-driven resources necessary to develop competitive models~\citep{tahir2025benchmarking,ahuja2023megaverse}. The homogeneity of existing datasets and evaluation standards exacerbates the underrepresentation of diverse linguistic and cultural contexts in modern LLMs, highlighting the urgent need for targeted efforts to bridge these gaps and promote inclusivity in multilingual AI development.

To address all these challenges, Alif-1.0-8B-Instruct model offers a promising solution to the limitations of conventional multilingual training approaches. By leveraging a modified self-instruct technique, this model incorporates a carefully curated Urdu dataset, specifically designed to enhance Urdu generation quality, bilingual translation, culturally aware understanding, and Urdu-native chain-of-thought based reasoning capabilities. This unique multilingual synthetic data distillation approach not only improves the model’s performance on Urdu and English tasks but also upholds ethical commitments to safety and cultural sensitivity~\citep{Mitchell2019Model}. Prior research has demonstrated that tailored datasets significantly enhance the effectiveness of language models, enabling deeper linguistic and cultural understanding~\citep{Kulkarni2023Towards}. By using a carefully curated Urdu dataset, Alif-1.0-8B-Instruct addresses persistent challenges in multilingual language modeling within constrained computational budgets~\citep{Husan2023Teachers}.

Alif-1.0-8B-Instruct demonstrates a significant leap in Urdu-specific task comprehension, outperforming leading multilingual LLMs. Its training pipeline follows a structured process: continued pretraining to reinforce foundational understanding, fine-tuning on the synthetic Urdu-Instruct dataset to enhance comprehension, incorporation of translated Urdu data for broader knowledge, and replayed English data to mitigate catastrophic forgetting. As a result, Alif-1.0-8B-Instruct, built upon the pretrained Meta Llama-3.1-8B base, demonstrates superior performance compared to Llama-3.1-8B-Instruct.~\citep{grattafiori2024llama3herdmodels} in Urdu-specific benchmarks while maintaining strong English fluency. It also outperforms prominent multilingual models such as Mistral-7B-Instruct-v0.3~\citep{jiang2023mistral7b}, Qwen-2.5-7B-Instruct~\citep{qwen2025qwen25technicalreport}, and Cohere-Aya-Expanse-8B~\citep{dang2024ayaexpansecombiningresearch}, all within an optimized training budget of less than \$100. 

\subsection{Contribution}
Our work introduces several key contributions to the development and fine-tuning of large language models, particularly focusing on multilingual and Urdu-specific capabilities:

\begin{itemize}
    \item Multilingual Urdu-English Model: We present Alif-1.0-8B-Instruct, a multilingual (Urdu-English) model that outperforms leading multilingual LLMs on Urdu-translated MGSM~\citep{shi2022language,cobbe2021gsm8k}, and Alpaca Eval~\citep{alpaca_eval,dubois2024length,dubois2023alpacafarm}, Dolly General QA~\citep{DatabricksBlog2023DollyV2},  benchmarks. 

    \item Modified Self-Instruct Technique: We introduce an enhanced self-instruct approach using diverse prompts and a global task pool. Each task is guided by unique prompts and seed values to capture cultural diversity, output structure, and task-specific nuances. A centralized task pool with human feedback ensures uniqueness and prevents redundancy. This scalable method improves instruction quality and can be adapted to other low-resource languages for broader NLP development.

    \begin{figure*}[t]
  \centering
  \includegraphics[width=0.7\textwidth]{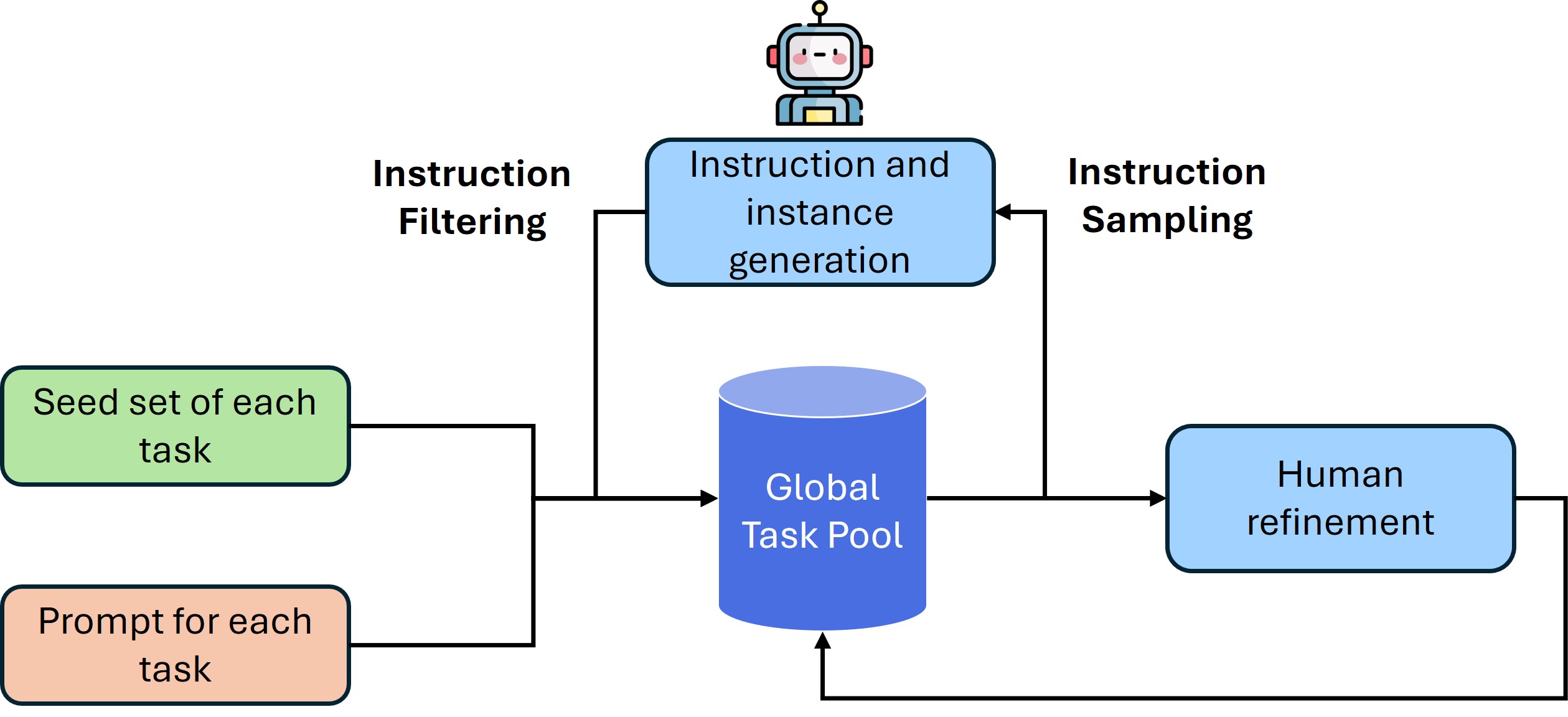}
  \caption{\label{fig-modified-self-instruct} Flowchart of the Modified Self-Instruct technique for Urdu-Instruct dataset generation.}
\end{figure*}

    \item High-quality Urdu-Instruct Dataset: We curated a high-quality multilingual synthetic dataset of 51,686 examples using a modified self-instruct method. It enriches Urdu capabilities through native chain-of-thought reasoning, bilingual translation, and cultural nuance. This approach also enabled the creation of a new Urdu evaluation set with $\sim$150 examples per task.

    \item Evaluations on Urdu-Translated Benchmarks and New Evaluation Dataset: We evaluate Alif-1.0-8B-Instruct on multiple Urdu-translated benchmarks, including MGSM, AlpacaEval, and Dolly General QA, demonstrating its effectiveness over state-of-the-art models. Results on our new Urdu evaluation set further highlight its strength in domain-specific tasks.

\end{itemize}


The rest of the paper is organized as follows: Section~\ref{sec:urdu-alpaca} introduces the Urdu-Instruct dataset and our modified self-instruct method. Section~\ref{sec:alif-model} details the Alif-1.0-8B-Instruct model, its training setup, and optimization techniques. Section~\ref{sec:results} presents evaluation results on Urdu and English tasks. Section~\ref{sec:perplexity} examines quantization impacts on performance and deployment.  Section~\ref{sec:conclusion} concludes with key takeaways and future directions in Urdu NLP and multilingual LLMs, followed by a discussion of the model’s limitations.


\section{Urdu-Instruct Dataset}
\label{sec:urdu-alpaca}

The Urdu-Instruct dataset, consisting of 51,686 examples generated using GPT-4o, api-version `2024-08-01-preview',~\citep{openai2024gpt4technicalreport}, is a crucial component in fine-tuning Alif-1.0-8B-Instruct. It contains instructions and responses for seven key Urdu tasks: Generation (5,907), Ethics (9,002), QA (8,177), Reasoning (9,590), Translation (10,001), Classification (4,662), and Sentiment Analysis (4,347). The dataset was created using a self-instruct~\citep{selfinstruct} technique improved for cultural and linguistic nuance as shown in Figure~\ref{fig-modified-self-instruct}\footnote{Bot image: \href{https://Flaticon.com}{Flaticon.com}} and explained below.
\subsection{Modified self-instruct technique}

\begin{enumerate}
\item Unique Prompt and Seed Values for each Task: To capture task-specific features, variations in output formats, and enhance cultural nuance, each task was assigned a distinct prompt and set of seed values. This ensured a richer and more diverse set of training examples, improving the model’s adaptability to different contexts.
\item Global Task Pool: While individual tasks had unique prompts and seed values, all generated instructions were consolidated within a single global task pool. This approach prevented duplication and ensured the uniqueness of each task distribution across the dataset. 
\item Instruction Sampling and Generation: Each prompt is augmented with random four human-annotated seed values and two machine-generated values to increase variability and ensure high-quality data. GPT-4o generates 20 instructions and corresponding outputs per batch.
\item Post-Processing and Filtering:
\begin{itemize}
\item Instructions shorter than three words or longer than 150 words were removed.
\item Instances containing unsuitable keywords for language models were filtered out.
\item Instructions starting with punctuation or containing characters other than Urdu and English, were rejected.
\item Each newly generated instruction was compared with all previously generated instructions across all tasks in the global task pool using a ROUGE score threshold of 0.7. Any instruction exceeding this similarity threshold was rejected.
\end{itemize}
\item Human Refinement: The dataset was further cleaned by human annotators to refine Urdu grammar, ensure factual correctness, and eliminate any accidental inclusion of unethical content or non-Urdu/non-English characters. Additional details are provided in \autoref{sec:appendixD}.
\end{enumerate}

\subsection{Urdu-Instruct dataset features}

This dataset covers a broad range of use cases, including text generation, ethical and safety considerations, factual question answering, logical reasoning, bilingual translation, classification, and sentiment analysis. Each task is designed to enhance the model's ability to understand and generate Urdu text effectively while maintaining high accuracy and cultural relevance.

\begin{itemize}
    \item CoT-Based Urdu Reasoning: We use Urdu-native Chain-of-Thought prompts and structured reasoning tasks to enhance the model’s logical abilities. This also improved performance in classification and sentiment analysis through better contextual understanding.
    \item Bilingual Translation: To reinforce the relationship between Urdu and English, we introduced bilingual translation tasks covering four distinct scenarios:
    \begin{table}[ht]
\centering

\begin{tabular}{lll}
\hline
\textbf{Instruction} & \textbf{Input} & \textbf{Output} \\
\hline
Urdu & English & Urdu \\
Urdu & Urdu & English \\
English & Urdu & English \\
English & English & Urdu \\
\hline
\end{tabular}
\caption{Instruction-Input-Output configurations.}
\label{tab:instruction-configs}
\end{table}

    \item Ethics and Safety: We align ethical considerations with cultural and regional norms, enabling more context-aware and safer AI behavior.
    \item Generation and QA: Incorporating both open- and closed-ended QA tasks improves Alif’s generation quality, coherence, and language understanding.
\end{itemize}

Using the same method, we created the Urdu Evaluation Set with $\sim$150 instructions per category, offering a benchmark for evaluating multilingual models on Urdu tasks.

\section{Multilingual Urdu-English Model: Alif-1.0-8B-Instruct} 
\label{sec:alif-model}

The development of Alif involves the integration of multiple datasets, each selected to serve a distinct role in the continued pre-training and fine-tuning process. This carefully structured approach is essential to enhancing the model's proficiency across a diverse range of tasks, ensuring robust linguistic capabilities.


\subsection{Datasets used for continued pre-training}\label{subsec:pretraining-datasets}

For the continued pre-training phase, we primarily utilize a dataset consisting of 200K Urdu Wikipedia articles\footnote{Dataset: \href{https://huggingface.co/datasets/wikimedia/wikipedia/viewer/20231101.ur}{wikimedia/wikipedia}}. This dataset is utilized to ensure diversity and coverage across multiple domains, aiming to provide a strong foundational understanding of language structures. By utilizing this dataset, we are able to maintain efficient training costs while ensuring the model achieved strong performance in text comprehension and generation tasks. We pre-train \textit{unsloth/Meta-Llama-3.1-8B}\footnote{Model:  \href{https://huggingface.co/unsloth/Meta-Llama-3.1-8B}{Meta-Llama-3.1-8B}} with the standard Causal Language Modeling (CLM) task. For an input tokens $\bm{x}=(x_0, x_1, x_2, \ldots)$, the model is trained to predict the next token as output $x_i$ autoregressively. The goal of the pre-training is to minimize negative log-likelihood loss as shown in equation~\ref{eq_pretrain}.
\begin{equation}
\label{eq_pretrain}
\mathcal{L}_{\textrm{CPT}}(\Theta) = \mathbb{E}_{\mathbf{x} \sim \mathcal{D}_{\textrm{PT}}} \left[ -\log p(\mathbf{x}; \Theta) \right]
\end{equation}

where $\Theta$ represents the model parameters, $\mathcal{D}_{\textrm{PT}}$ is the continued pre-training dataset, $x_i$ is the next token to be predicted, $x_0,x_1,\ldots,x_{i-1}$ is the input context, and CPT stands for continued pre-training.

\subsection{Datasets used for fine-tuning}\label{subsec:finetuning-datasets}

Alif is trained on a diverse collection of instruction-following datasets, comprising a total of 105,339 examples. These datasets include Urdu-Instruct (51,686 examples), translated dataset\footnote{Dataset: \href{https://huggingface.co/datasets/ravithejads/alpaca_urdu_cleaned_output}{ravithejads/alpaca\_urdu\_cleaned\_output}} (28,910 examples), ULS\_WSD (4,343 examples)~\citep{saeed2019word}, English Alpaca (10,400 examples)~\citep{english_alpaca}, and OpenOrca (10,000 examples)~\citep{OpenOrca, mukherjee2023orca, longpre2023flan, touvron2023llama2, touvron2023llama}.

The fine-tuning task is similar to the causal language modeling task: the model is prompted using the Stanford Alpaca template for fine-tuning and inference, and the input prompt looks like:

\begin{quote}\em\small
Below is an instruction that describes a task. Write a response that appropriately completes the request.
\newline
\newline
\#\#\# Instruction:
\newline
\{instruction\}
\newline
\newline
\#\#\# Input (If available):
\newline
\{input\}
\newline
\newline
\#\#\# Response: \{output\}
\end{quote}

The loss is only calculated on the \emph{\{output\}} part of the prompt and can be expressed as:
\begin{equation}
\label{eq_sft}
\mathcal{L}_{\textrm{SFT}}(\Theta) = \mathbb{E}_{\mathbf{x} \sim \mathcal{D}_{\textrm{SFT}}} \left[ -\log p(\mathbf{x}_{\textit{i}} \mid \mathbf{x}; \Theta) \right]
\end{equation}

Here, $\Theta$ represents the model parameters and  $\mathcal{D}_{\textrm{SFT}}$ is the fine-tuning dataset, $\bm{x}=(x_0, x_1, \ldots)$.


The selection of these datasets is strategically designed to strengthen the model’s instruction-following capabilities across multiple Urdu domains. Urdu-Instruct and translated datasets constitute the majority of the instruction-tuning data, while English Alpaca and OpenOrca are employed as replay datasets to mitigate catastrophic forgetting, preserving previously acquired knowledge throughout the fine-tuning process.


\subsection{Experimental setup and training details}

Low-Rank Adapters (LoRA) provide an efficient approach for continued pre-training and fine-tuning large language models, as introduced by \citep{hu2021loralowrankadaptationlarge}. This technique is particularly advantageous due to its computational efficiency, enabling model training without extensive GPU resources. We have employed LoRA and Unsloth framework\footnote{Website:  \href{https://unsloth.ai/}{unsloth.ai}} to optimize training costs while accelerating the overall training process. 
For our experiments, we utilized the \textit{unsloth/Meta-Llama-3.1-8B} as base model with LoRA applied to the following components:

\begin{itemize}
    \item QKVO (Self-Attention Layers): Query, Key, Value, Output projections.
    \item MLP (Feedforward Layers): Gate, Up, Down projections.
    \item ET-LH (Embedding \& Output Layers): Embedding tokens and Language Model Head.
\end{itemize}

By leveraging LoRA adapters, we have optimized the base model efficiently. The continued pre-training phase is conducted using Wikipedia articles, followed by fine-tuning. The training is performed using BF16 precision to ensure stability and efficiency. A cosine learning rate scheduler is employed, with an initial learning rate of $2 \times 10^{-5}$ for continued pre-training and $5 \times 10^{-5}$ for fine-tuning.

For training stage, we have utilized an Nvidia A100 GPU with 80GB of VRAM. The model is pre-trained for one epoch over 200K wikipedia dataset, requiring 23 hours on Runpod\footnote{Website:  \href{https://www.runpod.io/}{runpod.io}}. The fine-tuning phase, consisting of two epochs, have taken an additional 16 hours. We have accessed the A100 GPU via Runpod at a rate of \$1.64 per hour with a total training duration of 39 hours. As a result, the overall training cost remained under \$100 (as of February 12, 2025).

The detailed hyperparameters used for continued pre-training and fine-tuning are summarized in Table~\ref{pretraining-config}, with additional information provided in \autoref{sec:appendixC}.

\section{Results on Instruction-Following Tasks}
\label{sec:results}
Evaluating large language models (LLMs) for low-resource languages like Urdu presents unique challenges due to the limited availability of high-quality benchmarks. Additionally, while instruction-tuned models such as Llama-3.1-8B-Instruct have demonstrated strong multilingual capabilities, their performance in Urdu NLP tasks remains underexplored. In this section, we benchmark Alif-1.0-8B-Instruct (Alif) against Llama-3.1-8B-Instruct (Llama) and other LLMs using the alpaca chat template across various benchmarks. These evaluations were conducted on Runpod, using an A40 GPU with 48GB VRAM.

\subsection{Results on Urdu-translated benchmarks}

  
To ensure a rigorous and fair evaluation, we employ GPT-4o~\citep{openai2024gpt4technicalreport}, a LLM-as-a-judge scoring mechanism. Each response is assigned a 10-point score. To enhance the reliability of automated scoring, we refine GPT-4o's evaluation with human feedback. Our process involves continuous monitoring of GPT-4o's explanations across various evaluation tasks, enabling human feedback to identify inconsistencies and improve the evaluation prompt accordingly. This iterative refinement ensures greater accuracy and consistency in the evaluation of Urdu NLP models.

\begin{figure}[h]
  \centering
  \includegraphics[width=0.95\linewidth]{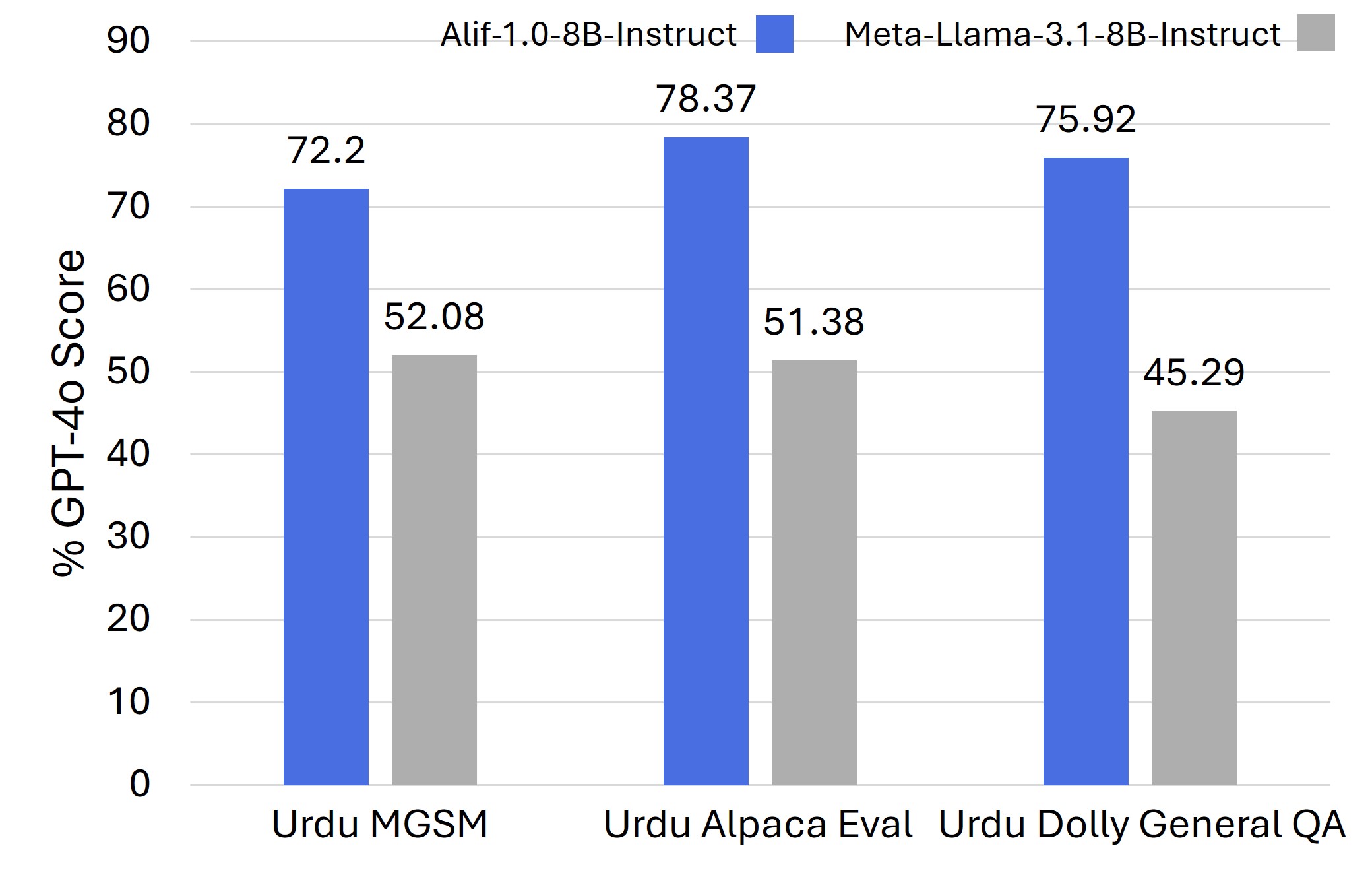}
  \caption{\label{fig-alif-meta} Comparison of Alif-1.0-8B-Instruct and Meta-Llama-3.1-8B-Instruct on Urdu-translated benchmarks.}
\end{figure}

\begin{table}[h]
  \centering
  \begin{tabular}{l c c}
    \hline
    \textbf{Task} & \textbf{Llama-3.1-Inst.} & \textbf{Alif-1.0-Inst.} \\
    \hline
    Generation & 42.8 & \textbf{90.2} \\
    Ethics & 27.3 & \textbf{85.7} \\
    QA & 30.5 & \textbf{73.8} \\
    Reasoning & 45.6 & \textbf{83.5} \\
    Translation & 58.9 & \textbf{89.3} \\
    Classification & 61.4 & \textbf{93.9} \\
    Sentiment & 54.3 & \textbf{94.3} \\
    \midrule
    Weighted Avg. & 45.7 & \textbf{87.1} \\
    \hline
  \end{tabular}
  \caption{Experimental results on Urdu evaluation set.}
  \label{table-alif-meta}
\end{table}

We utilize a structured prompt template to evaluate and compare the outputs of two systems, where System 1 represents the reference (ground-truth) response and System 2 is the generated response being evaluated. The model’s final score is computed as the percentage ratio of the System 2 score to the System 1 score, reflecting how closely the generated output aligns with the reference. The prompt template used for this evaluation is provided below. 

\begin{quote}\small\em 
You are an LLM Response Evaluator.\\

The following are two ChatGPT-like systems' outputs. Please evaluate both a ten-point scale (1–10), where 10 is the highest score, and provide a explanation for the scores. The evaluation criteria are:\\

- Relevance: Does the response directly and adequately address the user's prompt?\\
- Correctness: Is the information provided accurate and factually correct?\\
- Clarity: Is the response well-structured and free from unnecessary repetition or verbosity while maintaining completeness?\\
- Formatting Issues: Does the response have a consistent structure and free from unnecessary elements or incorrect language characters?\\

\#\#\# Prompt:
\{prompt\}\\
 
\#\#\# System1:
\{system1\_output\}\\

\#\#\# System2:
\{system2\_output\}
\end{quote}

We evaluate the models on a range of Urdu-translated benchmarks, including MGSM (250 math reasoning questions), AlpacaEval (806 instruction-following prompts), and a randomly sampled subset of Dolly General QA (220 open-ended questions). Across these diverse tasks, Alif consistently outperforms the base LLaMA model, demonstrating its improved reasoning and instruction-following capabilities in Urdu, as illustrated in Figure~\ref{fig-alif-meta}. 
Our evaluation also demonstrates that Alif significantly outperforms Llama in Urdu-specific NLP tasks, particularly in text generation, ethics, QA, translation, reasoning, classification, and sentiment as shown in Table~\ref{table-alif-meta}. 



\subsection{Results across different models}

\begin{table*}[t]
    \centering
    \begin{tabular}{l c  c c c}
        \hline
        \textbf{Models} & \textbf{MGSM} & \textbf{Alpaca Eval} & \textbf{Dolly General QA} & \textbf{Average} \\
        \hline  
        Falcon-7b-instruct & 21.0 & 23.2 & 21.4 & 21.8 \\
        Phi-3-small-8k-instruct & 43.1 & 38.7 & 35.6 & 39.1 \\
        Mistral-7B-Instruct-v0.3 & 43.6 & 43.6 & 38.7 & 41.9 \\
        Llama-3.1-8B-Instruct & 52.1 & 51.4 & 45.3 & 49.6 \\
        Granite-3.2-8b-instruct & 52.4 & 60.4 & 52.9 & 55.3 \\
        Gemma-7b-it & 57.5 & 58.0 & 54.5 & 56.6 \\
        Qwen2.5-7B-Instruct & 62.7 & 61.5 & 55.2 & 59.8 \\
        Ministral-8B-Instruct-2410 & 69.4 & 62.2 & 54.4 & 62.0 \\
        Aya-expanse-8b & 65.2 & 72.3 & 69.4 & 68.9 \\
        \textbf{Alif-1.0-8B-Instruct} & \textbf{72.2} & \textbf{78.4} & \textbf{75.9} & \textbf{75.5} \\
        \hline
    \end{tabular}
    \caption{Comparison of Alif-1.0-8B-Instruct with other models on Urdu translated benchmarks.}
    \label{tab:performance_other_models}
\end{table*}

Table \ref{tab:performance_other_models} presents a comparative evaluation of Alif-1.0-8B-Instruct against several leading instruction-tuned models on Urdu-translated benchmarks, including MGSM, Alpaca Eval, and Dolly General QA. The results indicate that Alif-1.0-8B-Instruct consistently outperforms all other models, achieving the highest scores across all three benchmarks. Specifically, it attains 72.2 on MGSM, 78.4 on Alpaca Eval, and 75.9 on Dolly General QA, leading to an overall average of 75.5. These results suggest that Alif-1.0-8B-Instruct is exceptionally well-suited for handling Urdu-based NLP tasks, demonstrating superior reasoning, comprehension, and instruction-following capabilities.


These results highlight the efficacy of Alif-1.0-8B-Instruct in tackling Urdu-translated benchmarks with a clear performance advantage over its counterparts. 

    

\subsection{Results on English benchmarks}

To assess whether Alif-1.0-8B-Instruct experiences catastrophic forgetting after adapting to Urdu, we evaluate its performance against Llama-3.1-8B-Instruct on a series of English-language benchmarks using \textit{lm-evaluation-harness}~\citep{lm-eval-harness} as shown in Table~\ref{tab:performance_metrics}. Since English data was incorporated during fine-tuning as a replay dataset, we anticipate that Alif-1.0-8B-Instruct should maintain competitive results on English tasks.

The evaluation results show that Alif-1.0-8B-Instruct retains strong general reasoning capabilities and even outperforms Llama-3.1-8B-Instruct in benchmarks such as \textit{arc\_challenge}, \textit{arc\_easy}, and \textit{hellaswag}, indicating that common sense and logical reasoning abilities are preserved. 

However, a slight decline is observed in knowledge-intensive tasks, particularly \textit{mmlu} where Llama-3.1-8B-Instruct achieves better results. The significant drop occurs in STEM and humanities categories of \textit{mmlu}, suggesting that while replay-based fine-tuning helps retain general capabilities, some domain-specific knowledge is affected.

Overall, these results indicate that using replay datasets during fine-tuning was effective in mitigating catastrophic forgetting, though some specialized knowledge areas experienced minor degradation. 

\begin{table*}[t]
    \centering
    \begin{tabular}{l c l c l c c}
        \hline
        \textbf{Tasks} & \textbf{Version} & \textbf{Filter} & \textbf{n-shot} & \textbf{Metric} & \textbf{Llama-3.1-Inst.} & \textbf{Alif-1.0-Inst.} \\
        \hline
        arc\_challenge & 1 & none & 0 & acc & 0.5171 & \textbf{0.5478} \\
                       &   & none & 0 & acc\_norm & 0.5512 & \textbf{0.5623} \\
        arc\_easy     & 1 & none & 0 & acc & 0.8190 & \textbf{0.8258} \\
                      &   & none & 0 & acc\_norm & 0.7950 & \textbf{0.8194} \\
        hellaswag    & 1 & none & 0 & acc & 0.5914 & \textbf{0.6135} \\
                     &   & none & 0 & acc\_norm & 0.7922 & \textbf{0.8022} \\
        mmlu         & 2 & none & 0 & acc & \textbf{0.6798} & 0.6177 \\
        \quad - humanities & 2 & none & 0 & acc & \textbf{0.6425} & 0.5530 \\
        \quad - other & 2 & none & 0 & acc & \textbf{0.7438} & 0.7007 \\
        \quad - social sciences & 2 & none & 0 & acc & \textbf{0.7702} & 0.7260 \\
        \quad - stem & 2 & none & 0 & acc & \textbf{0.5842} & 0.5268 \\
        \hline
    \end{tabular}
    \caption{Alif-1.0-8B-Instruct vs. Llama-3.1-8B-Instruct on English benchmarks.}
    \label{tab:performance_metrics}
\end{table*}

\section{Effect of Different Quantization Methods}
\label{sec:perplexity}
The deployment of large language models (LLMs) on various hardware architectures has traditionally been constrained by high computational and memory demands. However, the development of open source frameworks, such as \textit{llama.cpp}~\citep{llamacppGeorgi}, has facilitated the quantization of LLMs, significantly reducing their resource requirements and maintaining comparable accuracy for some quantized formats. This advancement also enables efficient local development, minimizing reliance on cloud services and enhancing data privacy. 

\subsection{Impact of quantization on Alif-1.0-8B-Instruct model}

\begin{figure}[ht]
  \centering
  \includegraphics[width=0.95\linewidth]{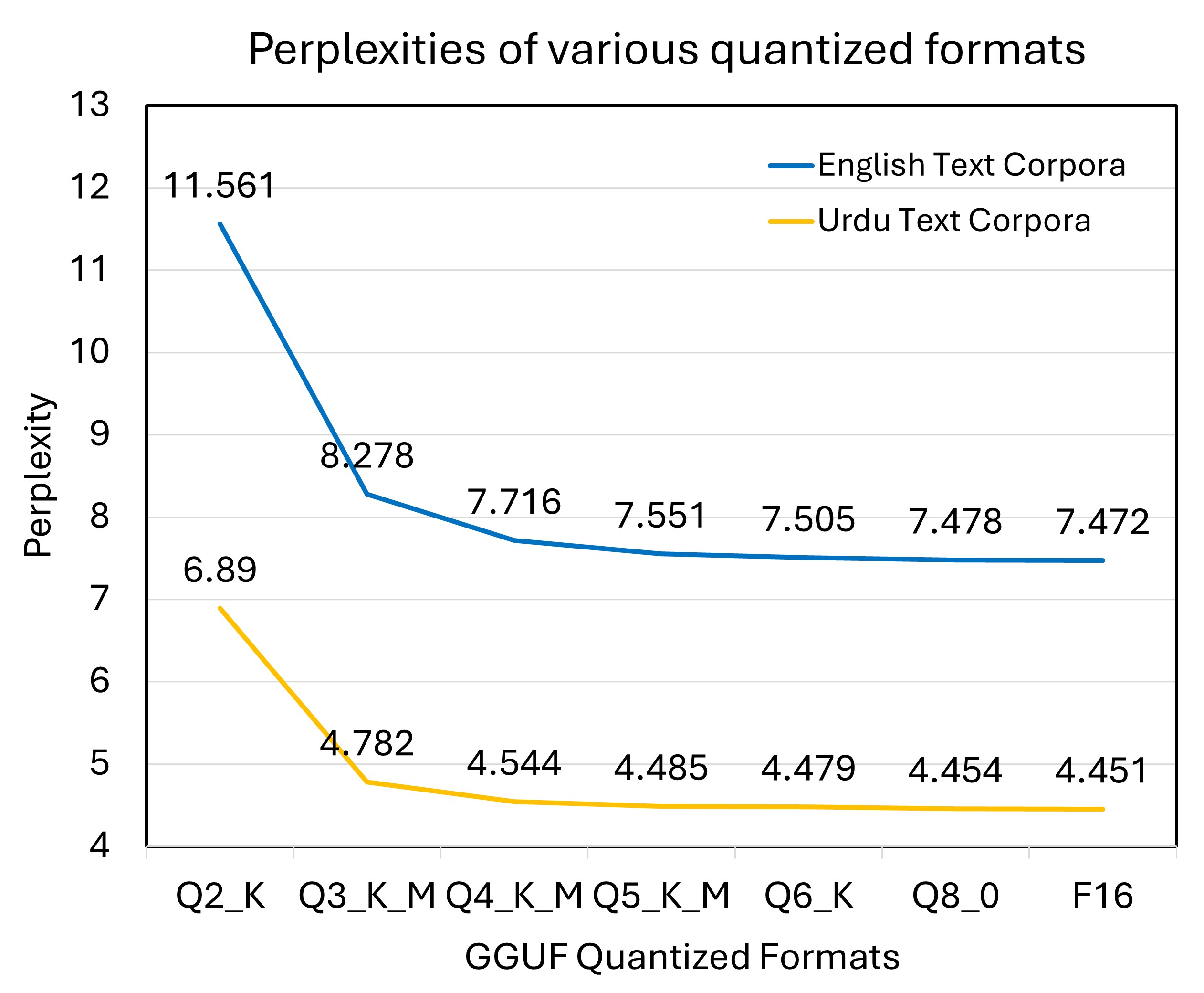}
  \captionsetup{justification=centerlast}
  \caption{\label{fig-perplexity1u} Perplexity comparison across GGUF quantization formats for Alif-1.0-8B-Instruct.}
\end{figure}

This section explores the effects of different quantizations on Alif-1.0-8B-Instruct model using \textit{llama.cpp}. We assess the model's perplexity (PPL) on English text corpora (wiki-test-raw) and a Urdu-translated version across various GGUF quantization formats.: \texttt{Q2\_K, Q3\_K\_M, Q4\_K\_M, Q5\_K\_M, Q6\_K, Q8\_0}, and \texttt{F16} (Half-precision). The results are depicted in Figure~\ref{fig-perplexity1u}.

\begin{figure}[ht]
  \centering
  \includegraphics[width=0.95\linewidth]{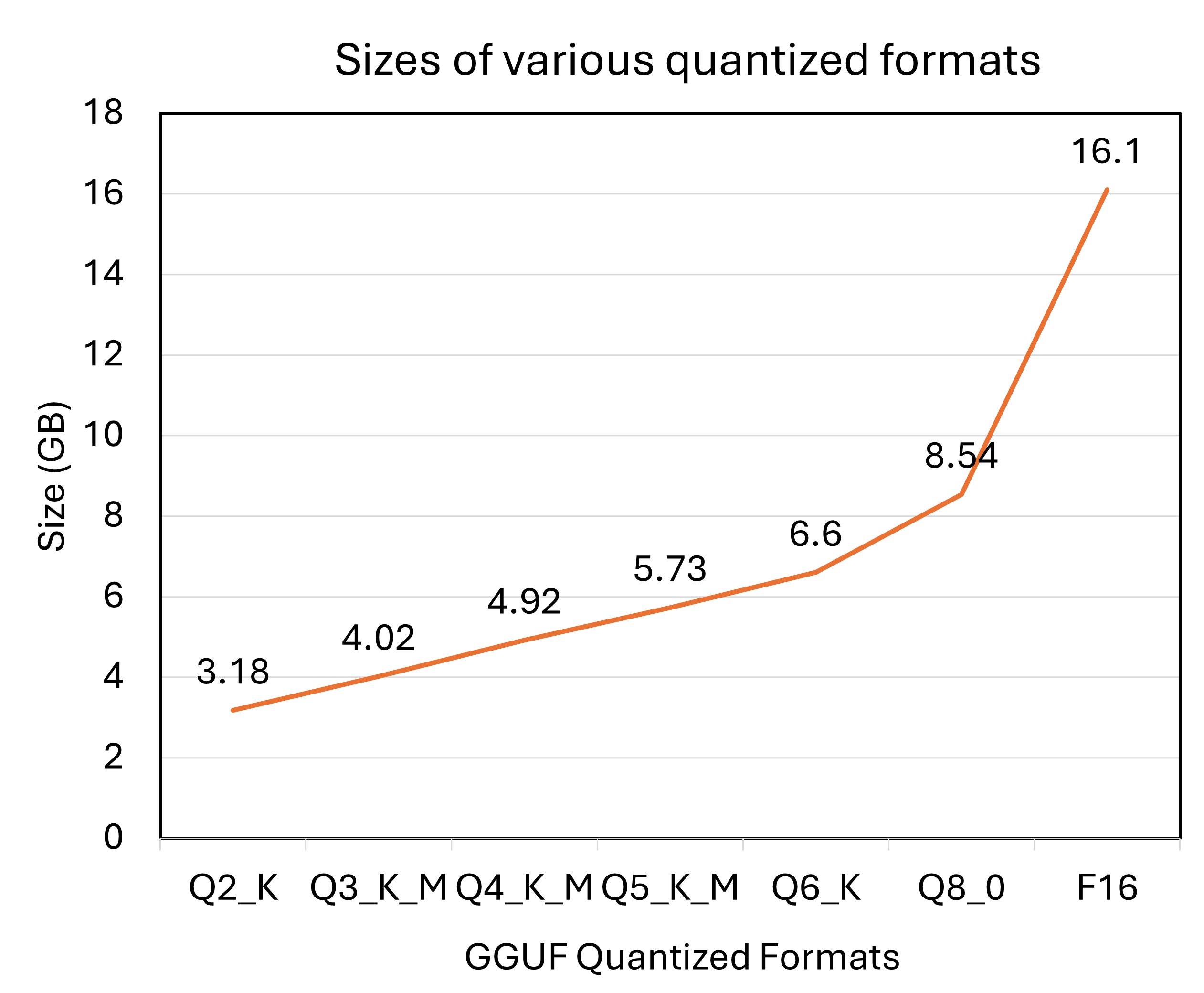}
  \captionsetup{justification=centerlast}
  \caption{\label{fig-perplexity2u} Memory footprint of different GGUF quantization formats for Alif-1.0-8B-Instruct.}
\end{figure}

Higher-bit quantization formats such as 6-bit and 8-bit maintain similar perplexity levels to FP16 while substantially reducing model size as shown in Figure~\ref{fig-perplexity2u}. Conversely, lower-bit quantization (2-bit, 3-bit, and 4-bit) results in higher perplexity, highlighting a tradeoff between efficiency and accuracy. The Urdu text corpus consistently shows lower perplexity compared to the English corpus, indicating better adaptation or linguistic properties influencing the model's comprehension.


Among the quantization format results, \texttt{Q6\_K} and \texttt{Q8\_0} emerge as optimal choices for deployment on personal computers, offering a practical balance between model size and accuracy. Lower-bit quantization (\texttt{Q3\_K\_M}, \texttt{Q4\_K\_M}) remains a viable option for resource-limited scenarios but comes with tradeoffs in model performance. In contrast, \texttt{Q2\_K} does not appear to be a viable solution due to a substantial increase in perplexity.

\section{Conclusion}
\label{sec:conclusion}
Building a high-performing Urdu LLM presents distinct challenges, including data scarcity, translation quality issues, and reasoning complexity. Existing methods often depend on large-scale translations, which degrade quality and raise data curation and training costs. We address this issue by continued pre-training and fine-tuning Alif-1.0-8B-Instruct on a high-quality multilingual synthetic dataset, Urdu-Instruct, which captures cultural nuances, enables bilingual knowledge transfer, and enhances reasoning abilities.

To further strengthen the study, future work will incorporate objective, task-specific metrics such as Exact Match, F1, BLEU, COMET, and BERTScore to more rigorously quantify alignment, factuality, and stylistic correctness in bilingual settings. Comparing multiple judge models and prompts will help evaluate robustness across cultural and linguistic variations.

Moving forward, we aim to broaden high-quality datasets, enhance reasoning through model merging and reinforcement learning, and benchmark Alif against evolving multilingual and reasoning standards. Alif marks a key step toward culturally aligned, reproducible, and cost-effective Urdu NLP, driving inclusive and trustworthy AI forward.

\section*{Limitations}
\label{sec:limitations}
The Alif-1.0-8B-Instruct model, introduced in this paper, marks a significant step in Urdu NLP. However, in the spirit of rigorous research, it is imperative to discuss the inherent limitations that accompany this model.

\begin{itemize}
    \item Urdu Task-Specific Knowledge: Despite high-quality pretraining and fine-tuning data, including the Urdu-Instruct dataset covering classification, reasoning, ethics, translation, and QA, some domain-specific and nuanced linguistic aspects remain underrepresented, limiting performance on culturally rich tasks.

    \item Harmful and Unpredictable Content: While designed to reject unethical prompts, the model may still produce harmful or misaligned outputs due to contextual limitations.

    \item Lack of Robustness: The model can behave inconsistently or illogically when faced with adversarial or rare inputs, highlighting the need for improved resilience.
\end{itemize}

Although some of these challenges can be mitigated in future iterations, we see this work as a crucial foundation that will drive further advancements in LLMs for Urdu and other low-resource languages.

\bibliographystyle{acl_natbib}

\appendix

\section{Potential Risks}
\label{sec:appendixA}

While Alif-1.0-8B-Instruct marks significant progress in Urdu NLP, several potential risks accompany its use:

\begin{itemize}

    \item Harmful and Biased Outputs: Despite safety training, the model may still produce harmful, racist, or discriminatory content, especially in response to ambiguous or adversarial prompts.

    \item Misuse in Unregulated Settings: The model could be used to generate propaganda, hate speech, or misinformation in settings where content moderation tools are limited or absent.

    \item Over-reliance Without Standard Benchmarks: The lack of strong Urdu evaluation datasets may lead users to place too much trust in the model, particularly in sensitive areas such as education, law, or public services.
\end{itemize}

\section{Experiments Setup}
\label{sec:appendixC}
\subsection{Training and evaluation environment}
All pretraining and fine-tuning experiments for Alif-1.0-8B-Instruct were performed on an NVIDIA A100 GPU (80GB) using Runpod cloud infrastructure. The experiments were run within a Docker container configured with Python 3.10 and a 200GB persistent volume for model checkpoints, datasets, and logs. Model training leveraged the \textit{Unsloth} framework, which enables efficient fine-tuning through low-rank adaptation (LoRA) and memory optimization techniques. Hyperparameter details of the experiment are given in Table~\ref{pretraining-config}

\begin{table}[h]
    \centering
    \begin{tabular}{ccc}
    \hline
         \bf Configurations& \bf Pre-training& \bf Fine-tuning\\ 
         \hline
         Training Data&  200K& 105K\\ 
         Epochs& 1& 2\\
         Batch Size&  64& 64\\
         Dropout&  0.01& 0\\
         LR&  2e-5& 5e-5\\
         LR\_Type&  Cosine& Cosine\\
         Max Length&  2048& 2048\\ 
         LoRA Rank&  128& 128\\ 
         LoRA Alpha&  32& 32\\
         LoRA Modules&  \makecell{QKVO, MLP, \\ ET-LH}&\makecell{QKVO, MLP, \\ ET-LH}\\
         \makecell{Trainable \\Params(\%)}&14.72\%&14.72\%\\
         \makecell{Training\\Precision}& BF16& BF16\\
         Training Time& 23 hours& 16 hours\\
    \hline
    \end{tabular}
    \caption{Training Hyperparameters.}
    \label{pretraining-config}
\end{table}

The environment was based on CUDA 12.2 and included the following key components:

\begin{itemize}
    \item Model Training Frameworks:
    \begin{itemize}
        \item \texttt{transformers==4.47.1}
        \item \texttt{trl==0.13.0}
        \item \texttt{peft==0.14.0}
        \item \texttt{accelerate==1.2.1}
        \item \texttt{unsloth @ 5dddf27}
    \end{itemize}

    \item Core PyTorch and CUDA Stack:
    \begin{itemize}
        \item \texttt{torch==2.5.1+cu121}
        \item \texttt{torchvision==0.20.1+cu121}
        \item \texttt{torchaudio==2.5.1+cu121}
        \item \texttt{bitsandbytes==0.45.0}
        \item \texttt{xformers==0.0.29.post1}
    \end{itemize}

    \item Data Handling and Processing:
    \begin{itemize}
        \item \texttt{datasets==3.2.0}
        \item \texttt{pandas==2.2.2}
        \item \texttt{tqdm==4.67.1}
        \item \texttt{scikit-learn==1.6.0}
        \item \texttt{libcudf-cu12}, \texttt{cupy-cuda12x}
    \end{itemize}

    \item Experiment Tracking and Logging:
    \begin{itemize}
        \item \texttt{wandb==0.19.1}
    \end{itemize}
\end{itemize}

All models were trained using mixed-precision settings with gradient accumulation to enable scalable fine-tuning under limited GPU memory constraints. The evaluation was conducted in the same software environment as training, with the only difference being the GPU. Specifically, all evaluations were performed on an NVIDIA A40 GPU (48GB) using Runpod cloud infrastructure.

\subsection{Modified Self-Instruct environment}

The Urdu-specific instruction dataset used in this work was generated using a modified version of the Self-Instruct framework. This version was adapted to improve cultural relevance, apply toxicity filtering, and refine prompt structures for Urdu. The generation pipeline integrates language model prompting, semantic filtering, and instruction post-processing.

\begin{itemize}
    \item Platform and Configuration:
    \begin{itemize}
        \item Language Model: \texttt{gpt-4o} via \texttt{AzureOpenAI} API.
        \item Python Version: 3.10
        \item Concurrency: Multiprocessing with \texttt{Pool} (24 CPUs).
    \end{itemize}

    \item Core Python Dependencies:
    \begin{itemize}
        \item \texttt{openai} — GPT-4o API integration.
        \item \texttt{rouge\_score} — Semantic similarity filtering via ROUGE-L.
        \item \texttt{numpy} — Batch operations and scoring computations.
        \item \texttt{LughaatNLP} — Urdu-specific lemmatization and tokenization.
        \item \texttt{tqdm}, \texttt{json}, \texttt{multiprocessing}, \texttt{re} — Preprocessing and utilities.
    \end{itemize}
\end{itemize}

\subsection{Urdu-Translation of Benchmarks}

To translate benchmark datasets into Urdu, we used GPT-4o (API version: 2024-08-01-preview) with the following prompt to ensure high-quality, fluent, and culturally appropriate translations:

\begin{quote}\em\small
You are an expert in Urdu linguistics and translation. Translate the following sentence into Urdu with accurate grammar, natural fluency, and cultural appropriateness. The output should be only translation with no additional word.

Sentence: \{sentence\}
\end{quote}

This prompt was used to translate all examples in the MGSM (250 math questions), AlpacaEval (806 instructions), and a 220-example subset of Dolly General QA into Urdu.

\section{Datasets Refinement}
\label{sec:appendixD}

To refine the Urdu-Instruct dataset and Urdu-translated instruction data, we employed a structured human annotator selection process focused on linguistic quality and demographic diversity.

\begin{figure}[ht]
  \centering
  \includegraphics[width=0.95\linewidth]{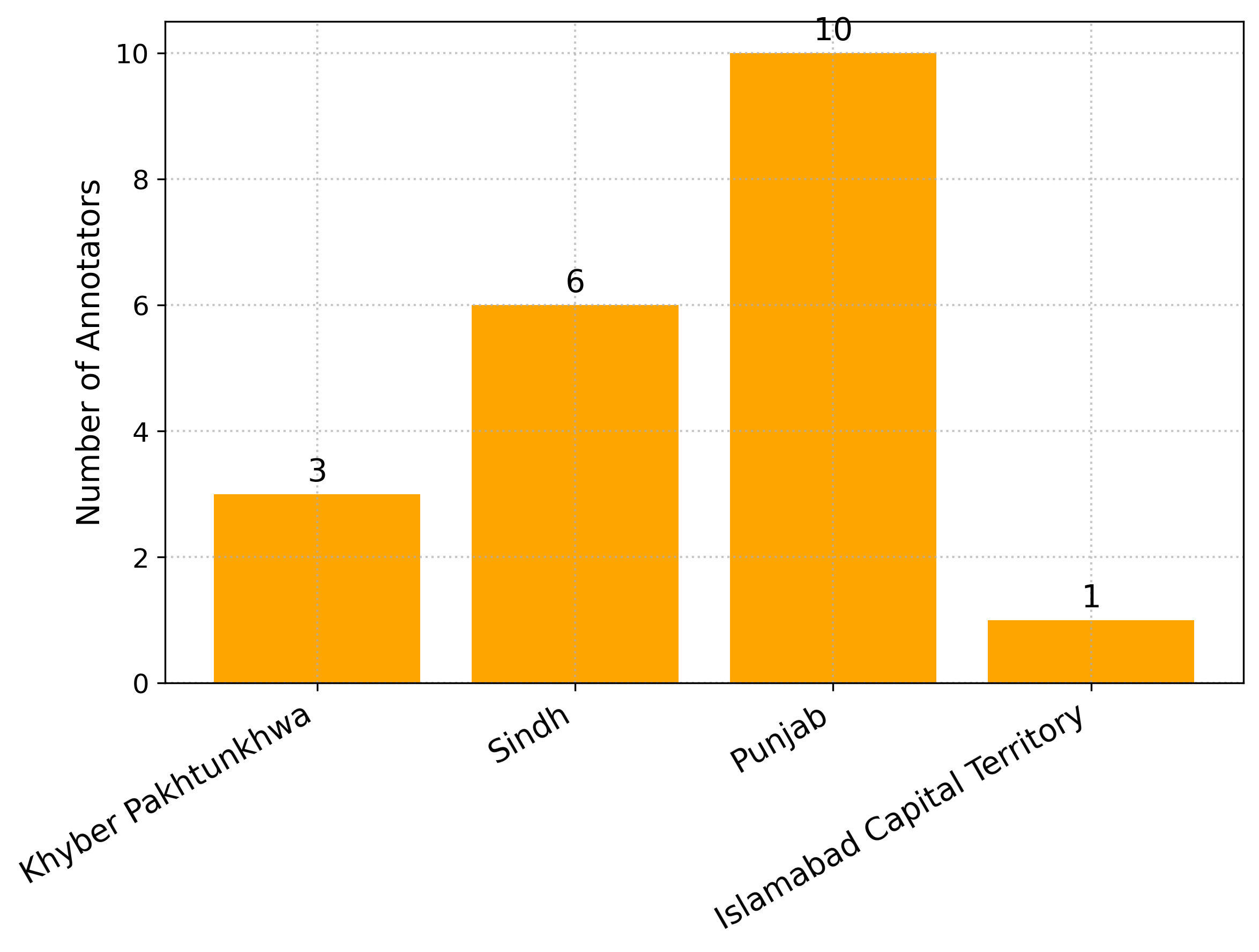}
  \captionsetup{justification=centerlast}
  \caption{\label{fig: demographics} Annotator Demographics by Province in Pakistan.}
\end{figure}

\begin{figure*}[t]
  \centering
  \includegraphics[width=0.95\textwidth]{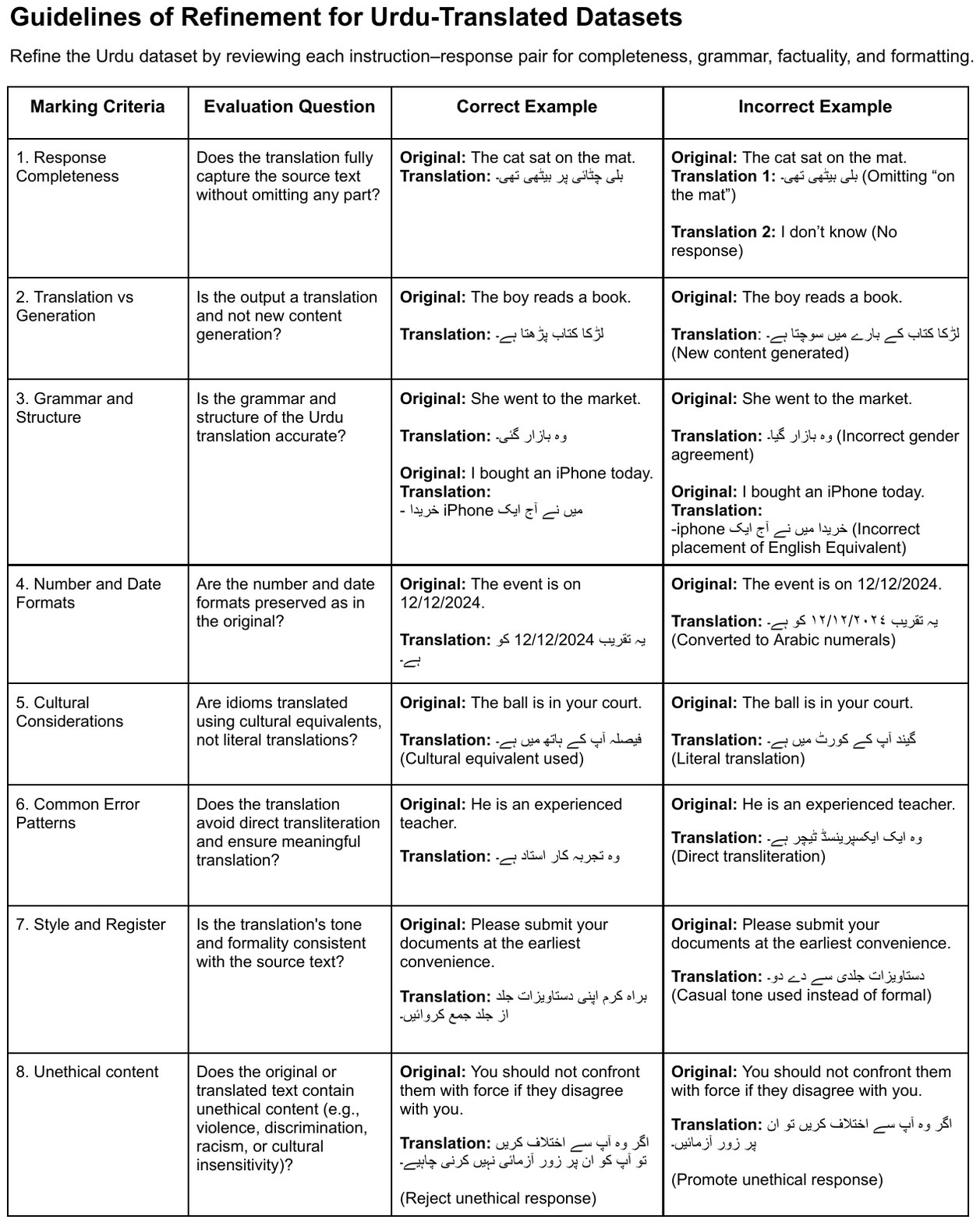}
  \caption{Overview of the Urdu-Translated datasets refinement guidelines.}
  \label{fig:refinement_urdutrans}
\end{figure*}

\begin{itemize}
    \item Recruitment:
    \begin{itemize}
        \item A public call for annotators was posted in October 2024, targeting native Urdu speakers with fluent typing skills and basic Excel knowledge.
        \item The opportunity offered task-based compensation, with applications collected via a form by October 16, 2024.
    \end{itemize}

    \item Shortlisting and Evaluation:
    \begin{itemize}
    \item Candidates were asked to complete two tasks: (1) correcting an error-filled Urdu passage, and (2) verifying and correcting 50 Urdu-translated instructions in an Excel sheet using the guideline as shown in Figure~\ref{fig:refinement_urdutrans}.
        \item 98 applicants attempted evaluation google form and 20 were shortlisted based on diverse demographics and task performance, and on-boarded to a dedicated Discord workspace. Among them, 14 were in the 18–24 age group, while 6 were between 25–34 years old and belong to various parts of Pakistan as shown in~Figure \ref{fig: demographics}.
    \end{itemize}

    \item Final Selection:
    \begin{itemize}
        \item These annotators, together with the author(s), contributed to refine translated and modified self-instruct datasets using the guidelines shown in~Figure \ref{fig:refinement_urduinstruct} and \ref{fig:refinement_urdutrans}.
        \item Annotators were compensated at a rate of 1000 Pakistani Rupees per hour, which is 4$\times$ of the minimum wage of Pakistan.
    \end{itemize}
\end{itemize}

\subsection{Unethical Content Rejection}

\begin{figure*}[t]
  \centering
  \includegraphics[width=0.95\textwidth]{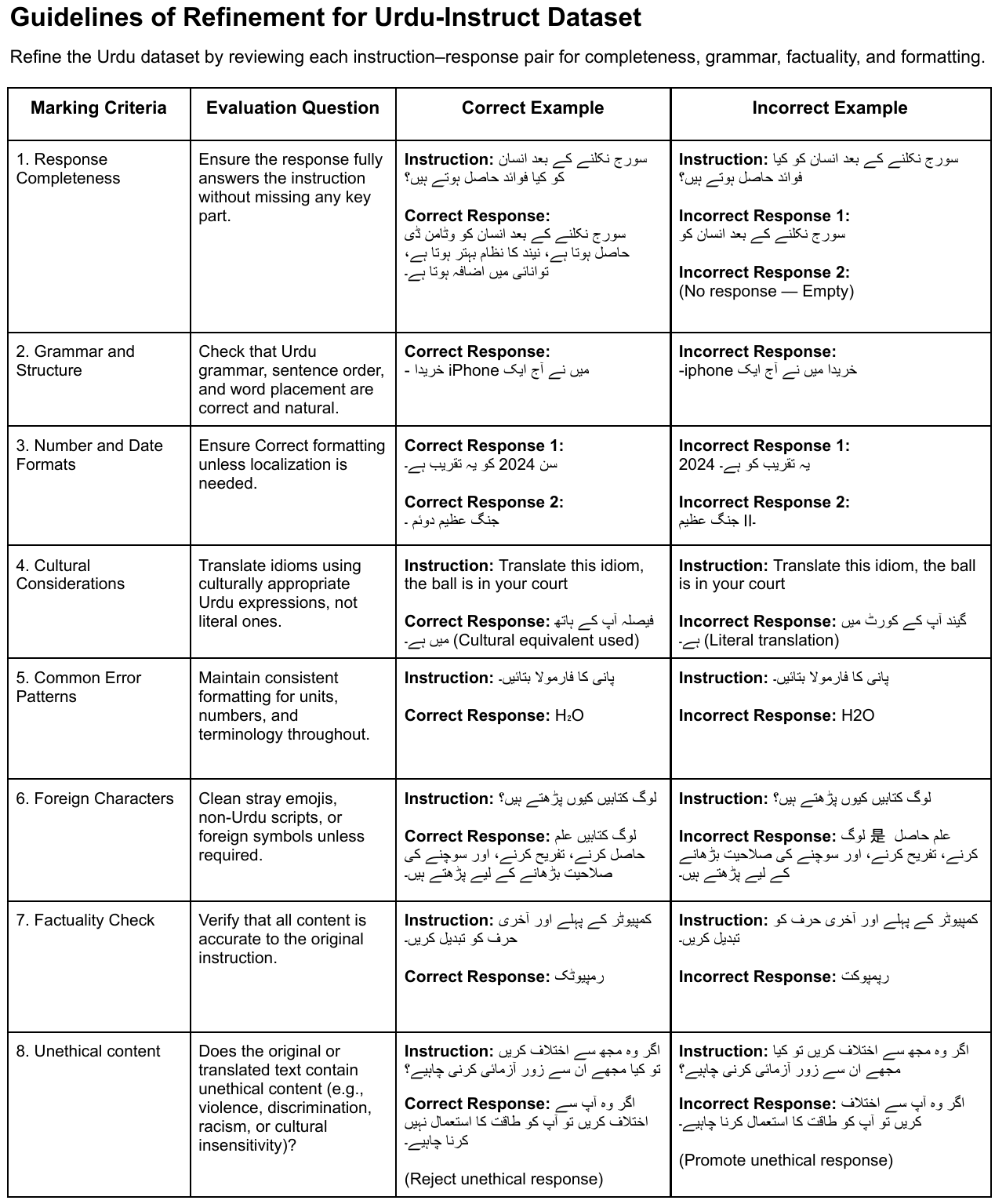}
  \caption{Overview of the Urdu-Instruct dataset refinement guidelines.}
  \label{fig:refinement_urduinstruct}
\end{figure*}

Unethical content was filtered out at two stages to ensure the quality and safety of the Urdu-Instruct dataset:

\begin{itemize}
    \item Automated Filtering: All translations and Urdu-Instruct generations were produced using GPT-4o via the Azure OpenAI API, which enforces strong safety guardrails to minimize the generation of harmful or inappropriate content.
    
    \item Human Refinement: A human annotation and review stage was conducted to further eliminate any accidental inclusion of unethical content, following a predefined set of refinement guidelines given in Figure~\ref{fig:refinement_urduinstruct} and \ref{fig:refinement_urdutrans}.
\end{itemize}

\section{AI Assistance}
 ChatGPT-4o model was used for fixing grammar issues, improving text readability, and coding support. All AI-generated content was reviewed and meticulously revised. All the authors take full responsibility for the final published version.

 \section{License}
\label{sec:license}

The Alif-1.0-8B-Instruct model is a continued pre-training and fine-tuning derivative of the Llama-3.1-8B base model, which is released under the \textit{Llama 3.1 Community License}.

The Urdu-Instruct dataset is released under the \textit{Creative Commons Attribution–ShareAlike 4.0 International (CC BY-SA 4.0)} License,\footnote{\url{https://creativecommons.org/licenses/by-sa/4.0/}} which allows use, modification, and redistribution with attribution, provided derivative works are shared under the same license. All source code used for data generation and fine-tuning is released under the \textit{MIT License}.

The datasets used for training the model are released under copyleft licenses, while the others are publicly available on Hugging Face without an explicitly specified license.

\end{document}